\documentclass{article} 
\usepackage{nips12submit_e,times}

\usepackage{times}
\usepackage{graphicx}
\usepackage{amsmath}
\usepackage{amssymb}
\usepackage{bm,amsfonts}
\usepackage{subfigure}
\newcommand\x{\mathbf{x}}

\newcommand\K{\mathbf{K}}

\renewcommand\c{\mathbf{c}}

\newcommand\C{\mathbf{C}}

\renewcommand\C{\mathbf{C}}
\newcommand\D{\mathbf{D}}
\newcommand\E{\mathbf{E}}
\newcommand\X{\mathbf{X}}
\newcommand\W{\mathbf{W}}

\newcommand\bLambda{\mathbf{\Lambda}}

\newcommand{\nystrom}{Nystr\"{o}m }

\newcommand{\hide}[1]{}

\title{Why Size Matters: \\Feature Coding as \nystrom Sampling}

\author{
Oriol Vinyals \\
UC Berkeley\\
Berkeley, CA \\
\And
Yangqing Jia \\
UC Berkeley\\
Berkeley, CA \\
\And
Trevor Darrell \\
UC Berkeley\\
Berkeley, CA \\
}
\nipsfinalcopy
\begin{document}
\maketitle


\section{Introduction}

Recently, the computer vision and machine learning community has been in favor of feature extraction pipelines that rely on a coding step followed by a linear classifier, due to their overall simplicity, well understood properties of linear classifiers, and their computational efficiency. In this paper we propose a novel view of this pipeline based on kernel methods and \nystrom sampling. In particular, we focus on the coding of a data point with a local representation based on a dictionary with fewer elements than the number of data points, and view it as an approximation to the actual function that would compute pair-wise similarity to all data points (often too many to compute in practice), followed by a \nystrom sampling step to select a subset of all data points.

Furthermore, since bounds are known on the approximation power of \nystrom sampling as a function of how many samples (i.e. dictionary size) we consider, we can derive bounds on the approximation of the exact (but expensive to compute) kernel matrix, and use it as a proxy to predict accuracy as a function of the dictionary size, which has been observed to increase but also to saturate as we increase its size. This model may help explaining the positive effect of the codebook size \cite{Coates2011ICML,yang2010efficient} and justifying the need to stack more layers (often referred to as deep learning), as flat models empirically saturate as we add more complexity.

\section{The \nystrom View}
We specifically consider forming a dictionary by sampling our training set. To encode a new sample $\x \in \mathbb{R}^d$, we apply a (generally non-linear) coding function $\c$ so that $\c(\x) \in \mathbb{R}^c$. Note that $d$ is the dimensionality of the original feature space, while $c$ is the dictionary size. The standard classification pipeline considers $\c(\x)$ as the new feature space, and typically uses a linear classifier on this space. For example, one may use the threshold encoding function \cite{Coates2011ICML} as an example: $\c(\x) = \max(0,\x^\top\D - \alpha)$ where $\D \in \mathbb{R}^{d \times c}$ is the dictionary. Note that our discussion on coding is valid for many different feed-forward coding schemes.

In the ideal case (infinite computation and memory), we encode each sample $\x$ using the whole training set $\X \in \mathbb{R}^{d \times N}$, which can be seen as the best local coding of the training set $\X$ (as long as over-fitting is handled by the classification algorithm). In general, larger dictionary sizes yield better performance assuming the linear classifier is well regularized, as it can be seen as a way to do manifold learning \cite{LCC}. We define the new coded feature space as $\C = \max(0,\X^\top\X - \alpha)$, where the $i$-th row of $\C$ corresponds to coding the $i$-th sample $\c(\x_i)$. The linear kernel function between samples $i$ and $j$ is $k(\x_i,\x_j) = \c(\x_i)^\top\c(\x_j)$. The kernel matrix is then $\K = \C\C^\top$. Naively applying \nystrom sampling to the matrix $\K$ does not save any computation, as every column of $\K$ requires computing an inner product with $N$ samples. However, if we decompose the matrix $\C$ with \nystrom sampling (i.e., with a subsampled dictionary) we obtain $\C' \approx \C$, and as a consequence $\K' \approx \K$:
$$\C' = \E \W^{-1} \E^\top, \quad %
\K' = \C' \C'^\top = \E \W^{-1} \E^\top \E \W^{-1} \E^\top = \E \bLambda \E^\top $$
where the first equation comes from applying \nystrom sampling to $\C$, $\E$ is a random subsample of the columns of $\C$, and $\W$ the corresponding square matrix with the same random subsample of both columns and rows of $\C$.

\section{Main Results on Approximation Bounds}

More interestingly, many bounds on the error made in estimating $\C$ by $\C'$ exist, and finding better sampling schemes that improve such bounds is an active topic in the machine learning community (see e.g. \cite{Kumar2012}). The bound we start with is \cite{Kumar2012}:

\begin{equation}
\label{eqn:nyst_bound}
||\C-\C'||_F \leq ||\C-\C_k||_F + \epsilon \max(n\C_{ii})
\end{equation}
valid if $c \geq 64k/\epsilon^4$ ($c$ is the number of columns that we sample from $\C$ to form $\E$, i.e. the codebook size), where $k$ is the sufficient rank to estimate the structure of $\C$, and $\C_k$ is the optimal rank $k$ approximation (given by Singular Value Decomposition (SVD), which we cannot compute in practice). Note that, if we assume that our training set can be explained by a manifold of dimension $k$ (i.e. the first term in the right hand side of eq. \ref{eqn:nyst_bound} vanishes), then the error is proportional to $\epsilon$ times a constant (that is dataset dependent).

Thus, if we fix $k$ to the value that retains enough energy from $\C$, we get a bound that for every $c$ (dimension of code), gives a minimum $\epsilon$ to plug in equation \ref{eqn:nyst_bound}. This gives us a useful bound of the form $\epsilon \geq M{c}^{-\frac{1}{4}}$ for some constant $M$ (that depends on $k$). Putting it all together, we get:
$$||\C-\C'||_F \leq O + M{c}^{-\frac{1}{4}}$$
with $O$ and $M$ constants that are dataset specific.

Having bounded the error $\C$ is not sufficient to establish how the code size will affect the classifier performance. In particular, it is not clear how the error on $\C$ affect the error on the kernel matrix $\K$. However, we are able to prove that the error bound on $\K'$ is in the same format as that on $\C$:
\begin{equation}
||\K-\K'||_F \leq O + M{c}^{-\frac{1}{4}}
\label{eqn:bound}
\end{equation}
Even though we are not aware of an easy way to formally link degradation in Frobenius norm of our approximation $\K'$ to $\K$ to classification accuracy, the bound above is informative as one may reasonably expect kernel matrices of different quality to have classification performances in the same trend.

\section{Experiments}

We empirically evaluate the bound on the kernel matrix, used as a proxy to model classification accuracy, which is the measure of interest. To estimate the constants in the bound, we do interpolation of the observed accuracy in the first two samples of accuracy versus codebook size, which is of practical interest: one may want to quickly run a new dataset through the pipeline with small codebook sizes, and then quickly estimate what the accuracy would be when running a full experiment with a much larger dictionary size.

\begin{figure}[th]
\centering
\begin{tabular}{cc}
  \includegraphics[width=0.35\linewidth]{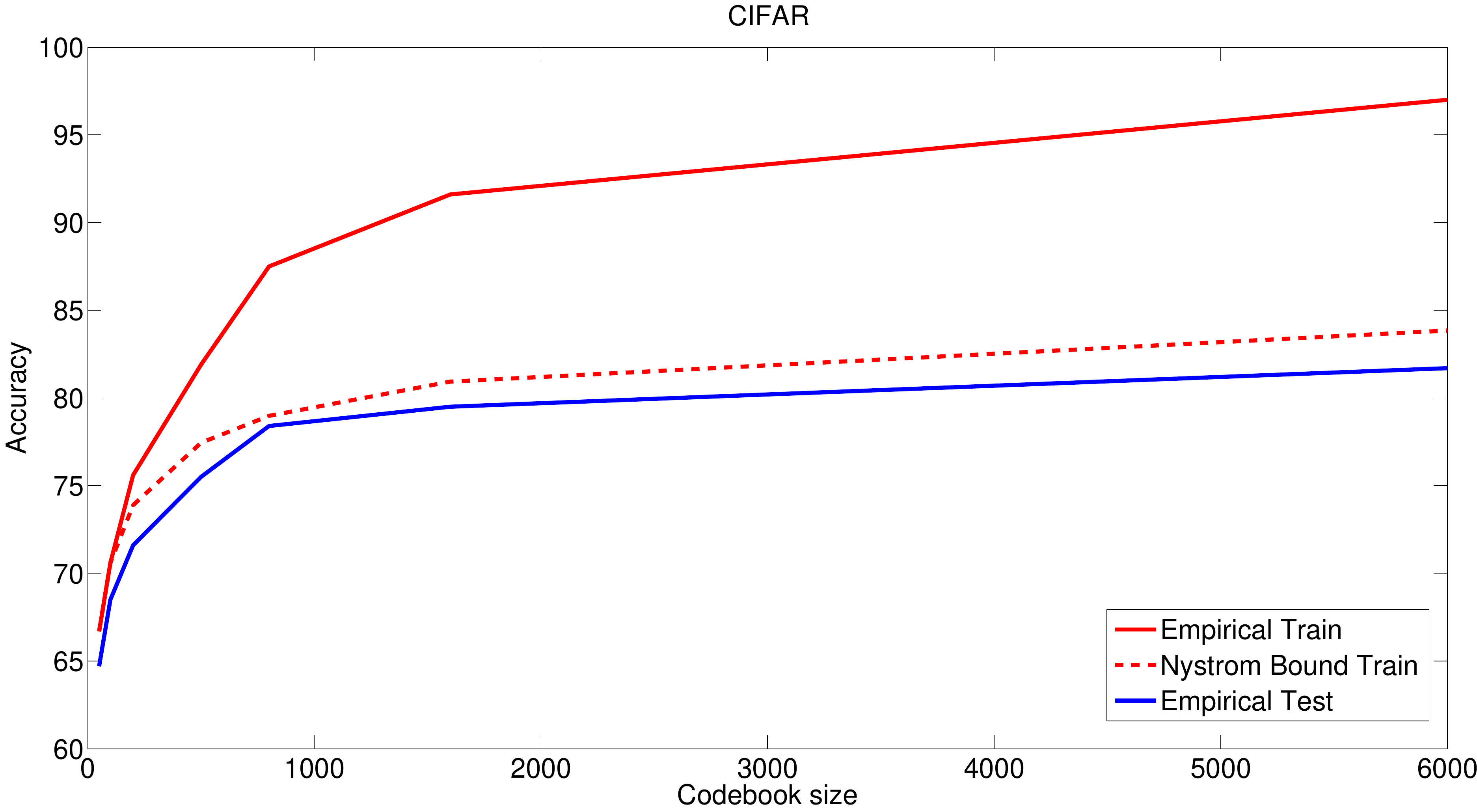} & 
  \includegraphics[width=0.35\linewidth]{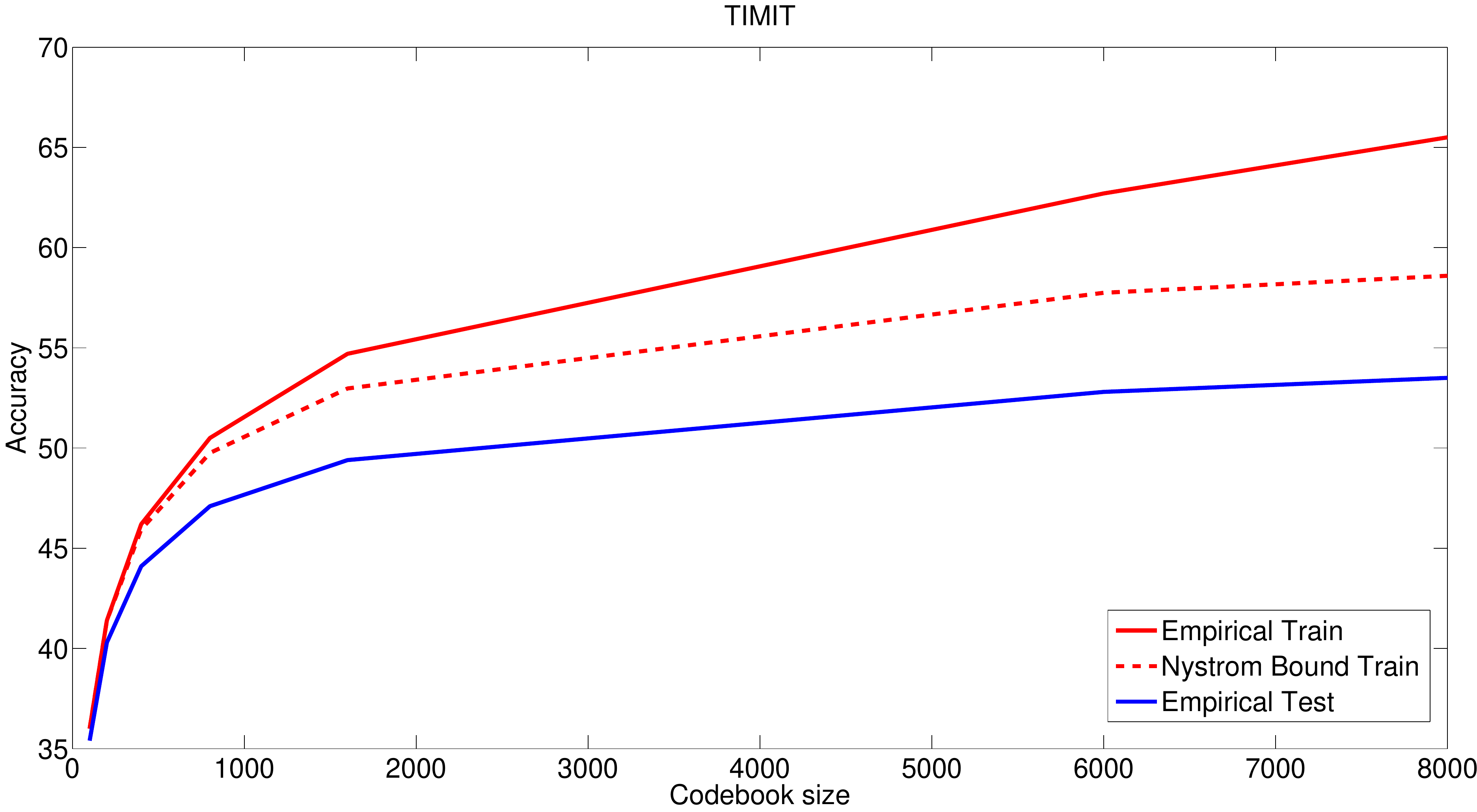} \\
\end{tabular}
\caption{Empirical accuracy (solid line) and \nystrom model accuracy (dashed line) on the training (red) and testing (blue) sets versus dictionary size, on CIFAR-10 (left) and TIMIT (right).}\label{fig:results}
\end{figure}

Figure \ref{fig:results} shows the results on on the CIFAR-10 image classification and TIMIT speech recognition datasets respectively. It is observed that the derived model closely follow our own empirical observations, with red dashed line serving as a lower bound of the actual accuracy and following the shape of the empirical accuracy, predicting its saturation. The model is never too tight though, due to various factors of our approximation, e.g., the analytical relationship between the approximation of $\K$ and the classification accuracy is not clear.

The \nystrom view of feature encoding and the approximation bounds we proposed helps understanding several key observations in the recent literature: (1) the linear classifier performance is always bounded when using a fixed codebook, and performance increases when the codebook grows \cite{Coates2011ICML}, even with a huge codebook \cite{yang2010efficient}, and (2) simple dictionary learning techniques have been found efficient in some classification pipelines \cite{coates2010aistats,saxe2011random}, and K-means works particularly well as a dictionary learning algorithm albeit its simplicity, a phenomenon that is common in the \nystrom sampling context \cite{Kumar2012}.

\begin{figure}
    \centering
    \includegraphics[width=0.35\textwidth]{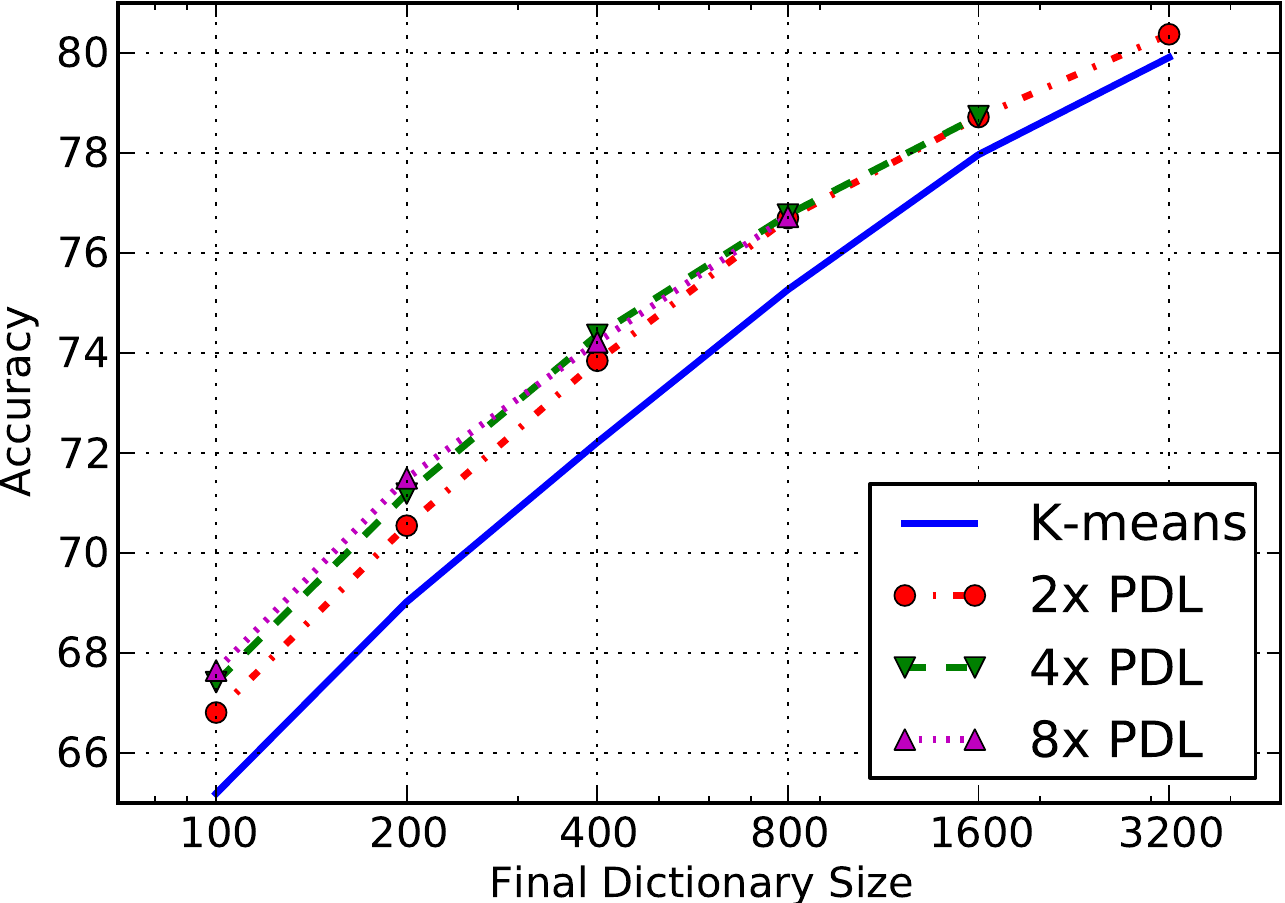}\hspace{0.1in}
    \includegraphics[width=0.35\textwidth]{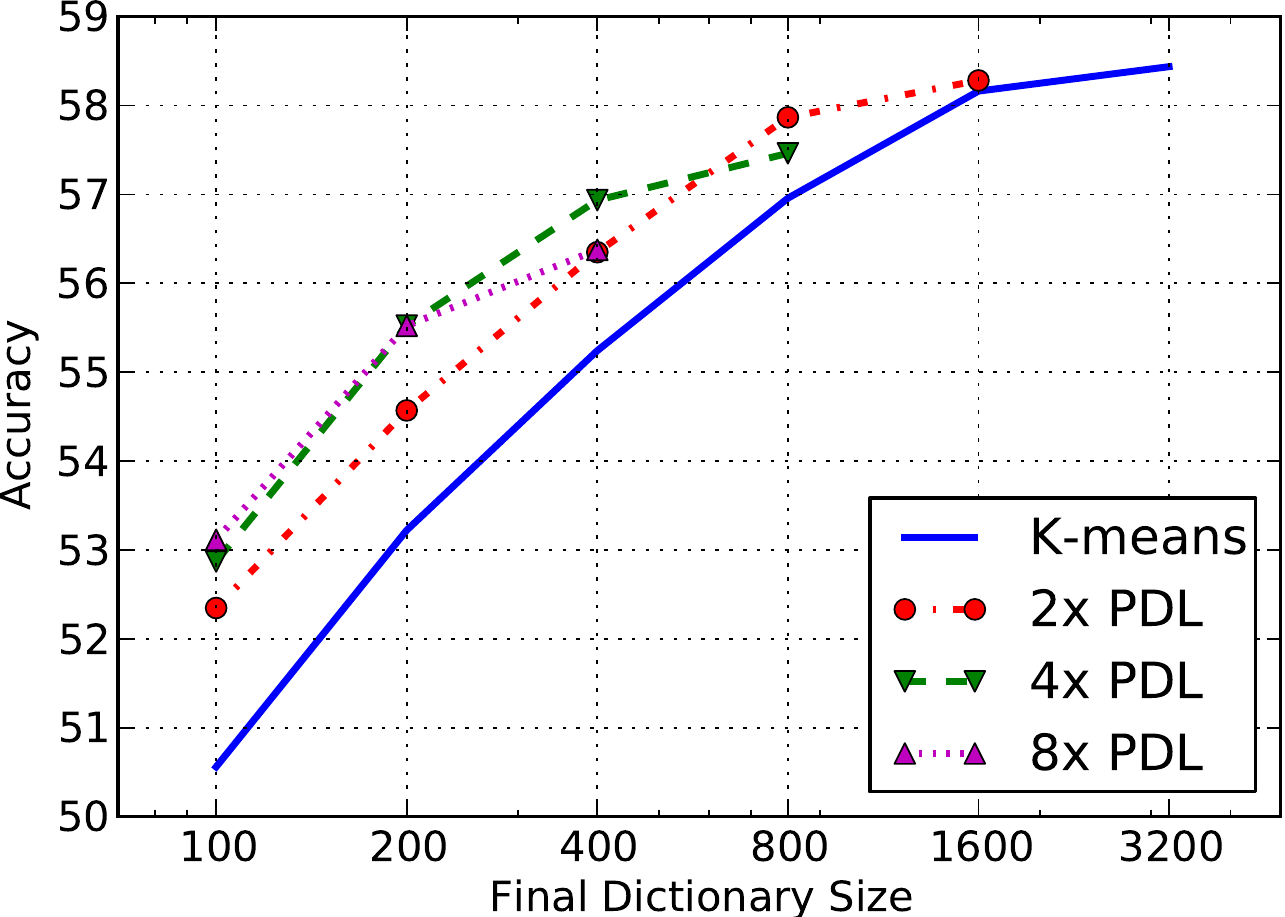}
    \caption{Accuracy values on the CIFAR-10 (left) and STL (right) datasets under different final dictionary size. ``nx PDL'' means overshooting the dictionary from a starting dictionary that is n times larger than the final one. We refer to our tech report \cite{jia2013pooling} for more details.}\label{fig:cifarstl}
\end{figure}

In addition, in many image classification tasks the feature extraction pipeline is composed of more than feature encoding. For example, recent state-of-the-art methods pool locally encoded features spatially to form the final feature vector. The \nystrom view presented in the paper inspires us to employ findings in the machine learning field to learn better, pooling-aware dictionaries. In one of our related work \cite{jia2013pooling}, we form a dictionary by first ``overshooting'' the coding stage with a larger dictionary, and then pruning it using K-centers with pooled features. Figure \ref{fig:cifarstl} shows an increase in the final classification accuracy compared with the baseline that only learns the dictionary on the patch-level, with no additional computation cost for either feature extraction or classification.

{
\bibliographystyle{plain}
\bibliography{refs}
}

\end{document}